\documentclass{article}

\usepackage{arxiv}
\usepackage{multicol}

\usepackage[utf8]{inputenc} 
\usepackage[T1]{fontenc}    
 \usepackage[colorlinks=true, citecolor=blue, urlcolor=blue, linkcolor=black]{hyperref} 
\usepackage{url}            
\usepackage{booktabs}       
\usepackage{amsfonts}       
\usepackage{nicefrac}       
\usepackage{microtype}      
\usepackage{lipsum}
\usepackage[english]{babel}
\usepackage{caption}
\usepackage{graphicx}
\usepackage{amsmath}

\makeatletter
\newenvironment{tablehere}
  {\smallskip\def\@captype{table}}
  {\smallskip}

\makeatother

\newcommand\fnurl[1]{%
\footnote{\url{#1}}%
}

\usepackage[sort,authoryear,round]{natbib}
\title{Face Detection with Feature Pyramids and Landmarks}

\author{\Large 
  Samuel W. F. Earp\thanks{Email: searp@sertiscorp.com}~, 
  Pavit Noinongyao, 
  Justin A. Cairns and 
  Ankush Ganguly \\ \\
  {\large Sertis Vision Lab}\thanks{597/5 Sukhumvit Road, Watthana, Bangkok, 10110, Thailand}~
}

\begin{document}
\maketitle


\begin{abstract}
Accurate face detection and facial landmark localization are crucial to any face recognition system.
We present a series of three single-stage RCNNs with different sized backbones (MobileNetV2-25, MobileNetV2-100, and ResNet101) and a six-layer feature pyramid trained exclusively on the WIDER FACE dataset.
We compare the face detection and landmark accuracies using eight context module architectures, four proposed by previous research and four modified versions.
We find no evidence that any of the proposed architectures significantly overperform and postulate that the random initialization of the additional layers is at least of equal importance.
To show this we present a model that achieves near state-of-the-art performance on WIDER FACE and also provides high accuracy landmarks with a simple context module.
We also present results using MobileNetV2 backbones, which achieve over $90\%$ average precision on the WIDER FACE hard validation set while being able to run in real-time.
By comparing to other authors, we show that our models exceed the state-of-the-art for similar-sized RCNNs and match the performance of much heavier networks.
\end{abstract}


\begin{multicols}{2}

\section{Introduction}


Over recent years, the ever-improving performance of Convolutional Neural Networks (CNNs) has resulted in highly accurate computer vision applications \citep[e.g. ][]{Redmon2018, Wang2018a, Tan2019, Wang2019, Wang2018b}.
One predominant application is object detection. 
The task consists of localizing objects and classifying them. 
Accurately solving this task opens the door to a wide range of applications, from autonomous driving \citep[see][for a review]{Badue2019} to person reidentification \citep[e.g.][]{hermans2017}.
The problem was traditionally approached in three main tracks, region selection \citep{Vedaldi2009}, feature extraction \citep{Dalal2005} and object classification \citep{Forsyth2014}.
However, the low efficiency of region selection and the semantic limitation of manually engineered descriptors hinders the performance of these algorithms.
Recently, many deep learning approaches have been developed to tackle this inefficiency issue. 
Region CNNs (RCNNs) \citep{girshick2014rich} are commonly used for various detection tasks. 
They perform a greedy selective search algorithm \citep{uijlings2013selective} to significantly lower the number of region propositions, however, this is computationally expensive.
Fast RCNN \citep{girshick2015} feeds pixel-level region proposals into the detection network from the feature maps, reducing the overhead somewhat.
However, Faster RCNNs \citep{ren2017}, utilize CNN based Region Proposal Networks (RPNs), removing the greedy selective search used in previous RCNNs, enabling detection in real-time.
RPNs use a pyramid of anchors to propose regions more efficiently than pyramids of images \citep[e.g.][]{viola2004} or filters. 
This anchor-based approach has successfully been applied many detection tasks \citep[][]{zhou2019,yang2018application,fan2016closer,sa2017intervertebral}.
Recently, single-stage detectors, based on Faster RCNN have also been widely adopted.
For example, \cite{lin2017} proposed a single-stage architecture called RetinaNet combined with focal loss---designed to combat the inherent class imbalance---which achieves state-of-the-art accuracy on the COCO dataset.

Face recognition is now commonplace in our daily lives, with businesses looking to take advantage of its convenience and robustness. 
Face detection serves as the foundation for recognition and various other face related research and products including alignment, and attribute classification (e.g. gender, age, face expression). 
Like any object detection task, the goal of face detection is to provide bounding boxes of all the faces in an image. 
However, variations in pose, illumination, resolution, occlusion, and human variance in real-world data make face detection challenging.
\cite{viola2004} proposed an approach that performs feature searching using Haar-like \citep{Wilson2006} features, which together with integral image, generates a set of features that assist in face detection. 
While the authors perform face detection on an image pyramid, the multi-scale feature pyramid approach has recently been shown to perform feature extraction more efficiently \citep{lin2017}.


Landmark localization refers to estimating predefined landmark locations in images.
Common tasks include facial (pupils, nose peak, mouth corners, etc.) and body (elbows, knees, wrists, shoulders, and face landmarks) landmark localization.
One of the most benchmarked public face landmark datasets is AFLW-2000 \citep{AFLW} which has 20,000 training images, 4386 test images and 19 manually annotated face landmarks per image.  
Facial landmarks are employed in various tasks, particularly in face alignment as a preprocessing step before face recognition \citep{taigman2014,sun2014}. 
One of the first breakthrough deep learning models for face landmark localization was Multitask Cascaded Convolutional Networks (MTCNN) \citep{zhang2016}. 
Utilizing several CNN networks, the authors achieved robust and accurate results and it has remained a strong baseline for several years.  
Subsequently, more computationally efficient methods have been developed which match the accuracy of MTCNN; such as \cite{bulat2017}.
Recently, methods superior to MTCNN--both in terms of accuracy and economy--have been developed including  RetinaFace \citep{deng2019}.
Many dense 3D face alignment techniques \citep[e.g.][]{Liu2018}, using U-Nets \citep{Guo2018} and Hourglass networks \citep{Newell2016} have further improved landmark localization.
However, these approaches often incur large computational overheads.

WIDER FACE\fnurl{http://shuoyang1213.me/WIDERFACE} \citep{yang2016} is a publicly available face detection benchmark dataset, which is widely used to train and benchmark face detection models.
The dataset contains 32,203 images and 393,703 faces selected from the publicly available WIDER dataset.
Recent work has produced very good results on this benchmark.
For example, Face Attention Networks (FAN) \citep{wang2017face} follows a similar approach to RetinaNet \citep{lin2017} using a single-stage and anchor-level attention networks trained in a supervised manner. 
They report an improvement over traditional methods and argue this is due to its capability to capture more contextual information.
Similarly, PyramidBox \citep{Tang2018} introduces a context-assisted single-stage detector that classifies and regresses faces, heads, and bodies to allow the detector to overcome small, blurred and partially occluded faces. 
The authors propose a data-anchor-sampling strategy which has subsequently been widely adopted  \citep{Li2019,zhang2019a,zhang2019c}.
PyramidBox++ \citep{Li2019} further enhances the original PyramidBox detector introducing progressive anchor loss \citep{Li2018} and by adding dense connections to the context module and employing a balanced-data-anchor-sampling strategy preventing oversampling on small faces.
RetinaFace \citep{deng2019} also use a RetinaNet approach and include landmarks, as well as, a 3D graph CNN mesh decoder alongside a joint shape and texture decoder \citep{zhou2019dense}, and a differentiable renderer \citep{Genova2018} to construct localized 3D face meshes.
This approach resulted in state-of-the-art performance on WIDER FACE.
AInnoFace \citep{zhang2019a} employs a slightly different strategy, using a modified RetinaNet to perform a two-stage classification and regression task. 
The authors, apply the Intersection over Union (IOU) regression loss to minimize the difference between predictions and ground-truths \citep{zhang2019a}, anchor-based sampling similar to the data-anchor-sampling in PyramidBox \citep{Tang2018}, and max-out \citep{zhang2017,Tang2018}.
The authors report similar results to RetinaFace on WIDER FACE.
RefineFace \citep{zhang2019b} achieves slightly higher performance on the WIDER FACE by combining five different modules; selective two-step regression, selective two-step classification, scale-aware margin loss, feature supervision module, and receptive field enhancement. 
The authors argue that these modules address the class imbalance, reduce the classifier search space and produce more discriminative features, and were able to get even better results on the WIDER FACE challenge.


This paper is organized as follows. 
Section 2 presents some of the most recent work that we will draw upon.
Section 3 details our loss function and context modules.
Section 4 outlines the experiment procedure we will employ.
Section 5 compares the results for different architectures.
Section 6 presents our conclusions.


\section{Related Work}


\subsection{Pyramid of features}
Recently the adoption of multi-scale feature pyramids for detection tasks has been widespread \citep[e.g. ][]{najibi2017,lin2016,zhang2017,wang2017face,deng2019,zhang2019a,zhang2019b}.
The work draws on results using spatial pyramid pooling \citep{he2014}, which can efficiently extract features at different levels from a single image, moving away from the less efficient pyramid of images approach \citep{viola2004}.
Multi-scale feature pyramids rely on only a single-scale image and outputs proportionally sized feature maps at various levels through top-down and lateral connections.
This approach displays significant performance improvements on the COCO  \citep{lin2014,lin2016}, and WIDER FACE \citep{wang2017face,deng2019,zhang2019a,zhang2019b} challenges.
Due to the performance of this approach, we will be adopting it in our experiments.


\subsection{Single versus two-stage}

Generally, there are two types of modern face detectors, single-stage, and two-stage.
Single-stage models make independent object classification from multiple feature maps from deep in the network \citep{Liu2016}, typically having a latency advantage.
However, these feature maps have a lower spatial resolution, hence may have already lost some semantic information relating to small objects, generally leading to reduced accuracy.
Two-stage detectors (e.g. Faster-RCNN) construct semantically rich feature maps from different layers in the network \citep{lin2017} and classify regions of interest. 
As a result, two-stage based architectures can detect small objects with higher precisions but with reduced speeds \citep{yoo2019}.
Finding a balance between accuracy and inference time has been a predominant focus of recent research.

\subsubsection*{Two-stage}
In the past few years, there have been numerous two-stage detectors that perform well on WIDER FACE.
For example, \cite{li2017} proposed `Light-Head' RCNNs, an efficient and accurate two-stage face detector by generating `thin' feature maps, applying a large-kernel deformable convolution before the RoI warping, inspired by Light RCNN. 
The authors add additional small anchors to support tiny faces which help in evaluation achieving $95.9\%$, $94.5\%$ and $87.9\%$ on the easy, medium and hard WIDER FACE sets.
Similarly, \cite{Li2018} proposed Duel Shot Face Detectors (DSFDs) using progressive anchor loss, a feature enhancement module and an improved anchor matching strategy to achieve state-of-the-art face detection. 
The authors use a Feature Enhance Module, a combination of a typical FPN and a Receptive Field Block 
(RFB) \citep{liu2017}, before the second shot.
They also propose a Progressive Anchor Loss strategy, using smaller anchors in the first shot and larger in the second, arguing that original feature maps have less semantic information but more location information.

\subsubsection*{Single-stage}
More recently, however, single-stage solutions have shown their dominance.
For example, \cite{najibi2017} introduced a Single Stage Headless (SSH) architecture that detects faces in a single forward pass by directly extracting features from different scales within the network.
Their detector achieved state-of-the-art performance on the WIDER FACE dataset and is eight times faster than previous methods.
More recently, RetinaFace \citep{deng2019}, another example of a RetinaNet \citep{lin2017} style single-stage feature pyramid detection network achieved state-of-the-art on WIDER FACE.
The authors stress the importance of incorporating face key-points with the bounding boxes for improved performance on WIDER FACE. 
Recent results have shown that single-stage detectors can outperform two-stage both in terms of accuracy and latency.
This work was subsequently followed by both \cite{zhang2019b} and \cite{zhang2019a} using similar approaches.
Because of this recent success, we will also be utilizing a single-stage RetinaNet approach.


\subsection{Multi-task learning}
\cite{Chen2014} were the first to propose combining face detection and alignment into a joint cascade framework.
Subsequently, other authors have used this approach to improve the accuracy of face detection networks \citep[e.g.]{chen2016,zhang2016,deng2019}.
Having a face detector that can also provide basic alignment information is extremely beneficial for any face recognition system.
Multi-task learning is a common practice for training face detection networks.
For example, \cite{Tian2018} presented a feature fusion pyramid architecture with a weakly supervised segmentation branch able to achieve state-of-the-art performance on WIDER FACE.
The authors used the combination of three loss functions to train their network; a classification loss, a regression loss, and a segmentation loss.
The authors argue that the segmentation branch helps the network learn more discriminative features. 
Similarly, \cite{deng2019} trained RetinaFace using both a face landmark loss and a dense regression loss---generated from the difference between the original face and the reconstruction from a mesh decoder.
Like \cite{Tian2018}, the authors were able to achieve state-of-the-art performance on WIDER FACE.


\subsection{Landmark localization}
Face recognition models rely on having well aligned faces at training and inference \citep{taigman2014,sun2014,deng2019}.
To align the face, a transformation on the original image is needed, such that the landmarks of each face should reside in a specific location.
This transformation depends on the quality of the landmark locations.
MTCNN \citep{zhang2016} has been used prolifically in face recognition tasks because the network provides both face bounding boxes and landmarks.
\cite{deng2019} improved on this with RetinaFace, a Faster-RCNN face detector that also returns the same face landmarks as MTCNN, so it can easily replace MTCNN in most use cases.
The authors found vast improvements in verification accuracy on LFW \citep{lfw}, CFP-FP \citep{cfp-fp}, AgeDB-30 \citep{agedb} and IJB-C \citep{ijbc} just by changing from MTCNN to RetinaFace.
Therefore, we will also include a landmark loss term to help train our networks.


\subsection{Context}
\cite{najibi2017} were the first to propose using context modules in single-stage detectors.
Since this work, several papers have used different context modules in their detectors and have reported improved results on WIDER FACE \citep[e.g.][]{deng2019,Li2019}.
In the original paper, SSH, the authors took the network output and performed a series of three $3 \times 3$ convolutions, then concatenated the outputs of the final two.
RetinaFace \citep{deng2019}, uses a similar 3 layer approach, reducing the number of filters in the second and third layers by a factor of two, then concatenates all three outputs.
However, in their GitHub repository\fnurl{https://github.com/deepinsight/insightface/tree/master/RetinaFace} they also use a second context module which sums the output of the first two layers and concatenates it with the first.
\cite{Li2019} use densely connected convolutions \citep{huang2016} in their context module, where the input of each convolutional layer is the concatenation of all previous layers.
In this paper, we will be comparing all of these context modules, alongside a few of our own, to quantify the influence of the context module's architecture on the overall performance of the network.




\section{PyramidKey}


\subsection{Multi-task loss function}
As previously mentioned using a multi-task loss function is commonplace in detection tasks.
To train our models we use a loss function comprised of three components; class, bounding box and landmark loss.
The class loss $L_{\text{cls}}$ is given by the log loss over the two classes (face vs background), this is calculated for both positive and negative anchors.
The bounding box loss $L_{\text{box}}$ is the smooth-$L_1$ regression loss of the box location, only calculated for positive anchors.
Similarly, the landmark loss $L_{\text{lmk}}$ is the regression loss of the landmark locations, also only calculated for the positive anchors.
The combination of these three loss functions yields our multi-task loss function,
\begin{align}
\begin{split}
&L(\{p_i\}, \{t_j\}, \{l_j\}) = \frac{1}{N_{\text{cls}}} \sum_i L_{\text{cls}}(p_i, p^*_i) \\
&+ \frac{1}{N_{\text{reg}}} \sum_j \big[ \lambda_1 L_{\text{box}}(t_j, t^*_j) + \lambda_2 L_{\text{lmk}}(l_j, l^*_j) \big], 
\end{split}
\end{align}
%
where $\lambda_{1,2}$ are scale factors which are set to $0.25$ and $0.1$, $i$ and $j$ correspond to all and positive anchors, $N_{\text{cls}}$ and $N_{\text{reg}}$ are the batch size and number of anchors, respectively.
\cite{lin2017} proposed using focal loss to address the inherent class imbalance, however, we find no significant benefit in replacing cross entropy.

\subsection{Context modules}
\begin{figure*}[t]
\centering
\includegraphics[height=0.15\textheight]{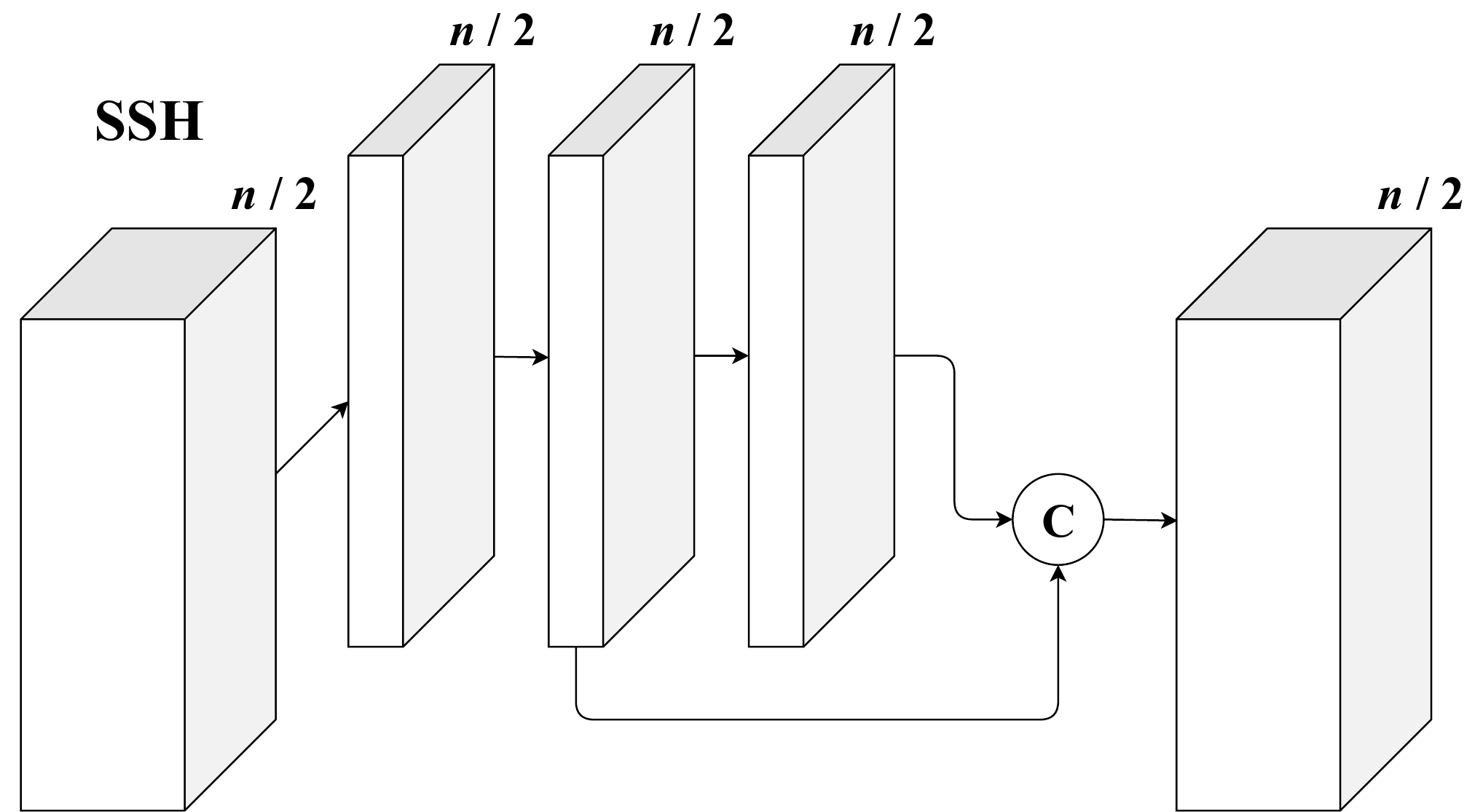}
\includegraphics[height=0.15\textheight]{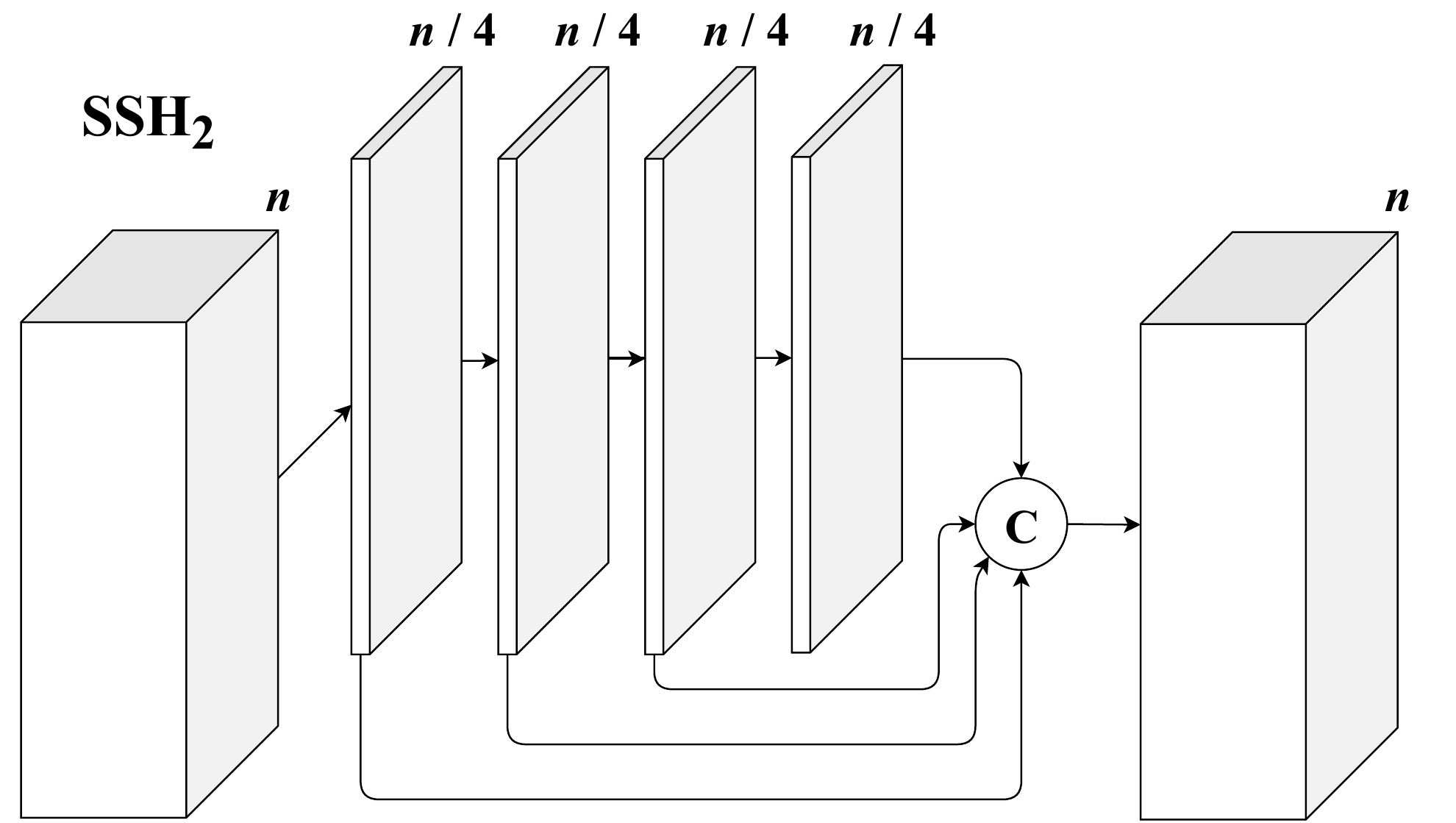}\\
\vskip 0.25cm
\includegraphics[height=0.15\textheight]{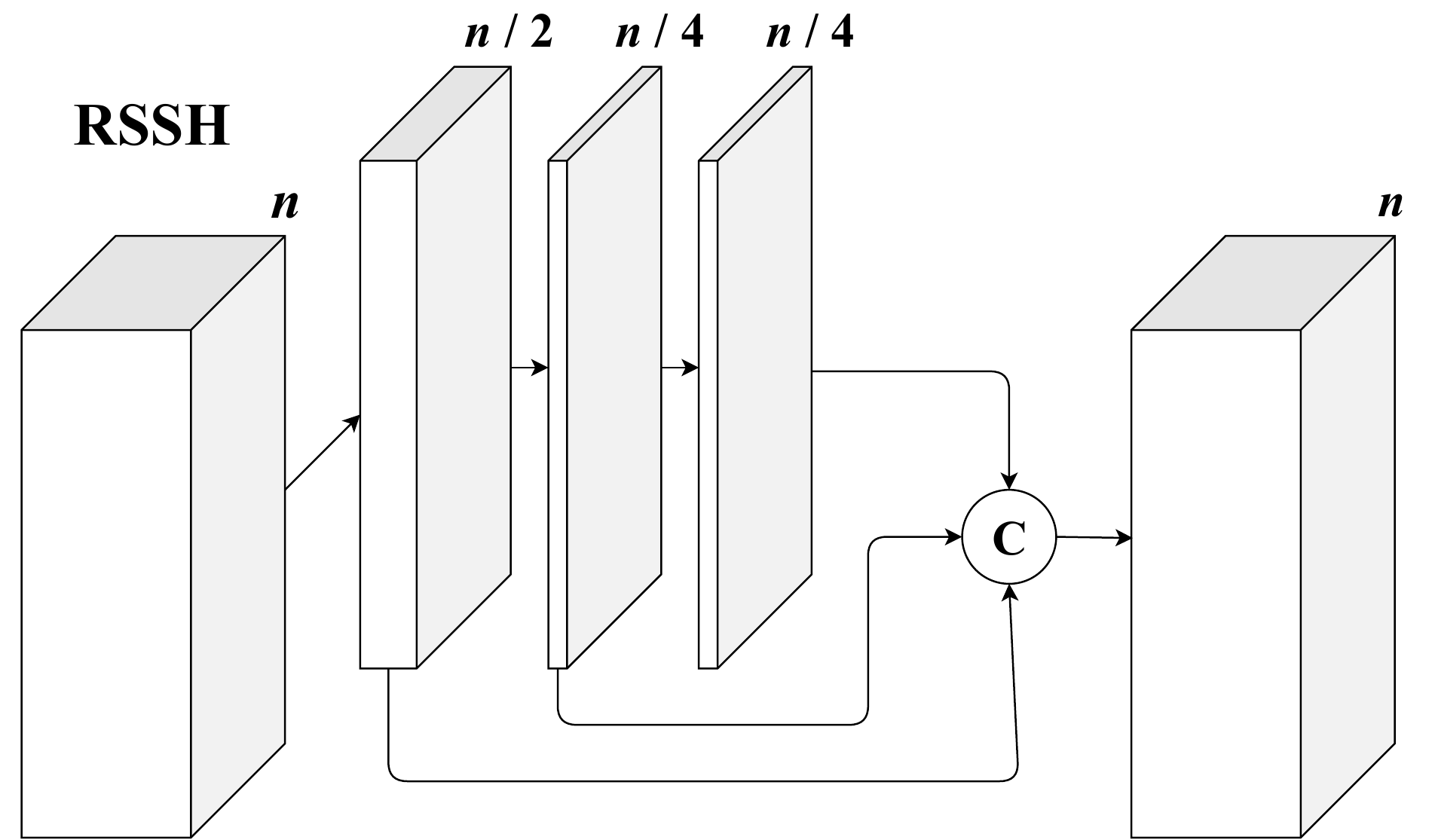}
\includegraphics[height=0.15\textheight]{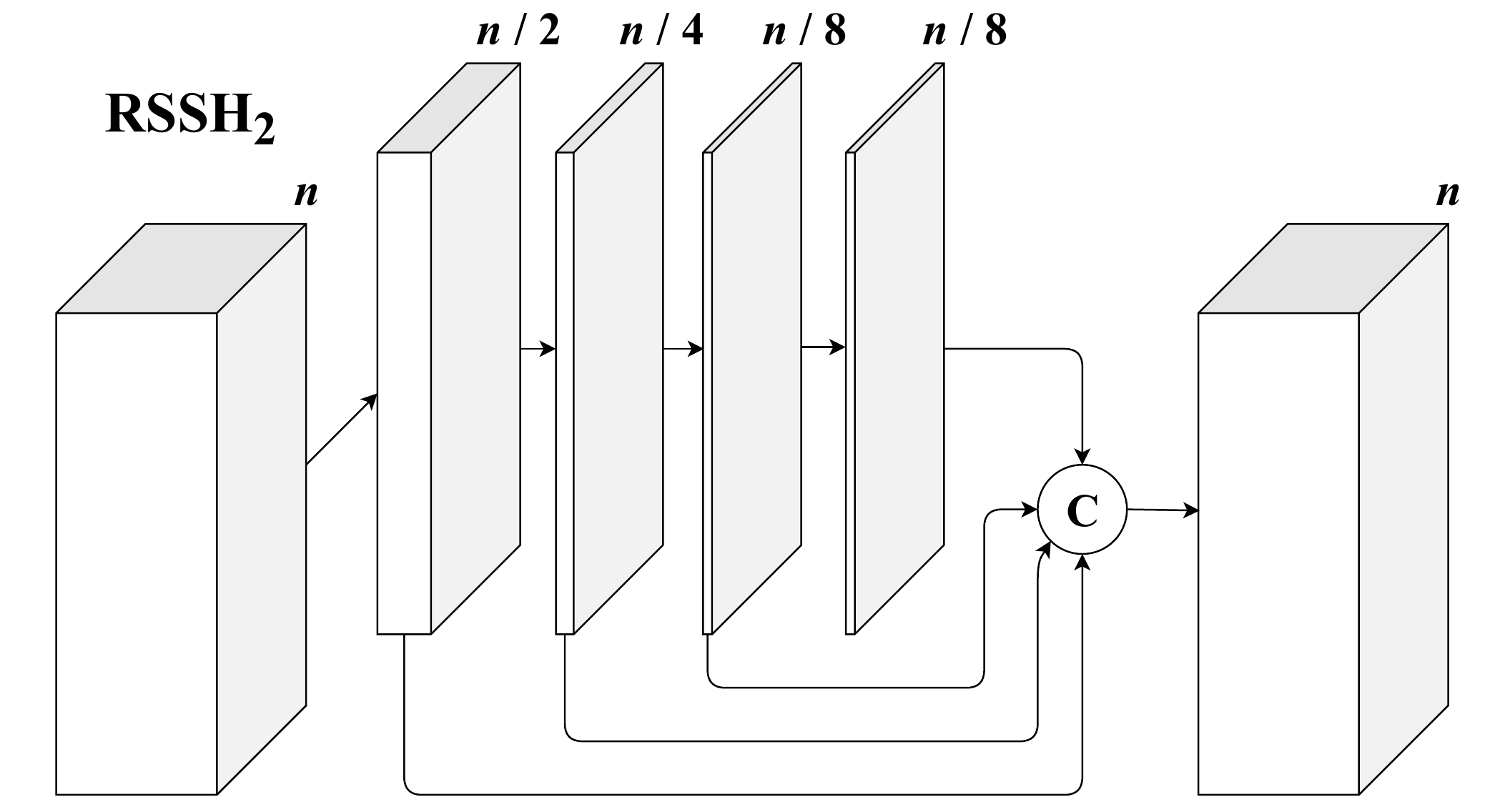}\\
\vskip 0.25cm
\includegraphics[height=0.15\textheight]{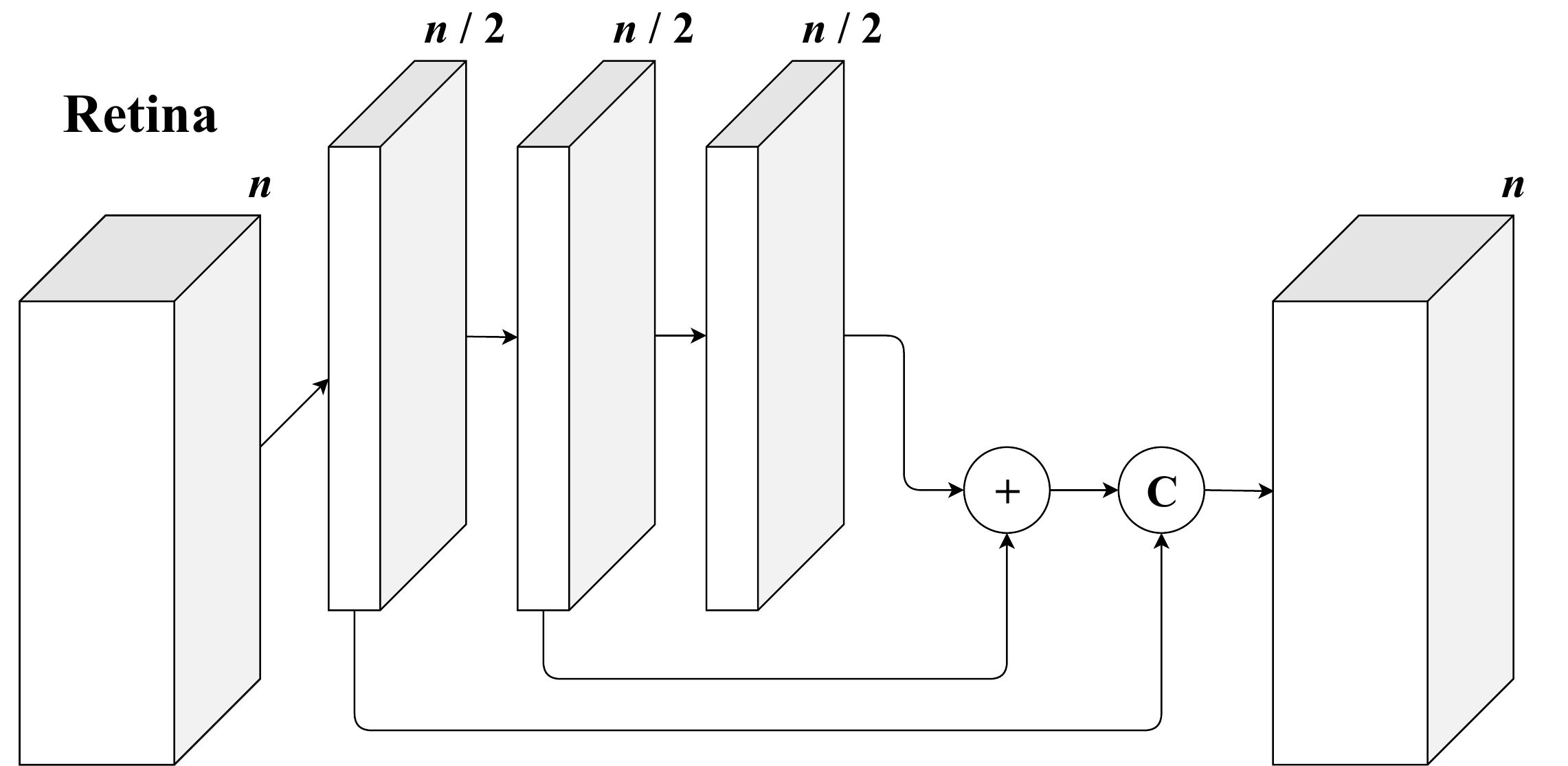}
\includegraphics[height=0.15\textheight]{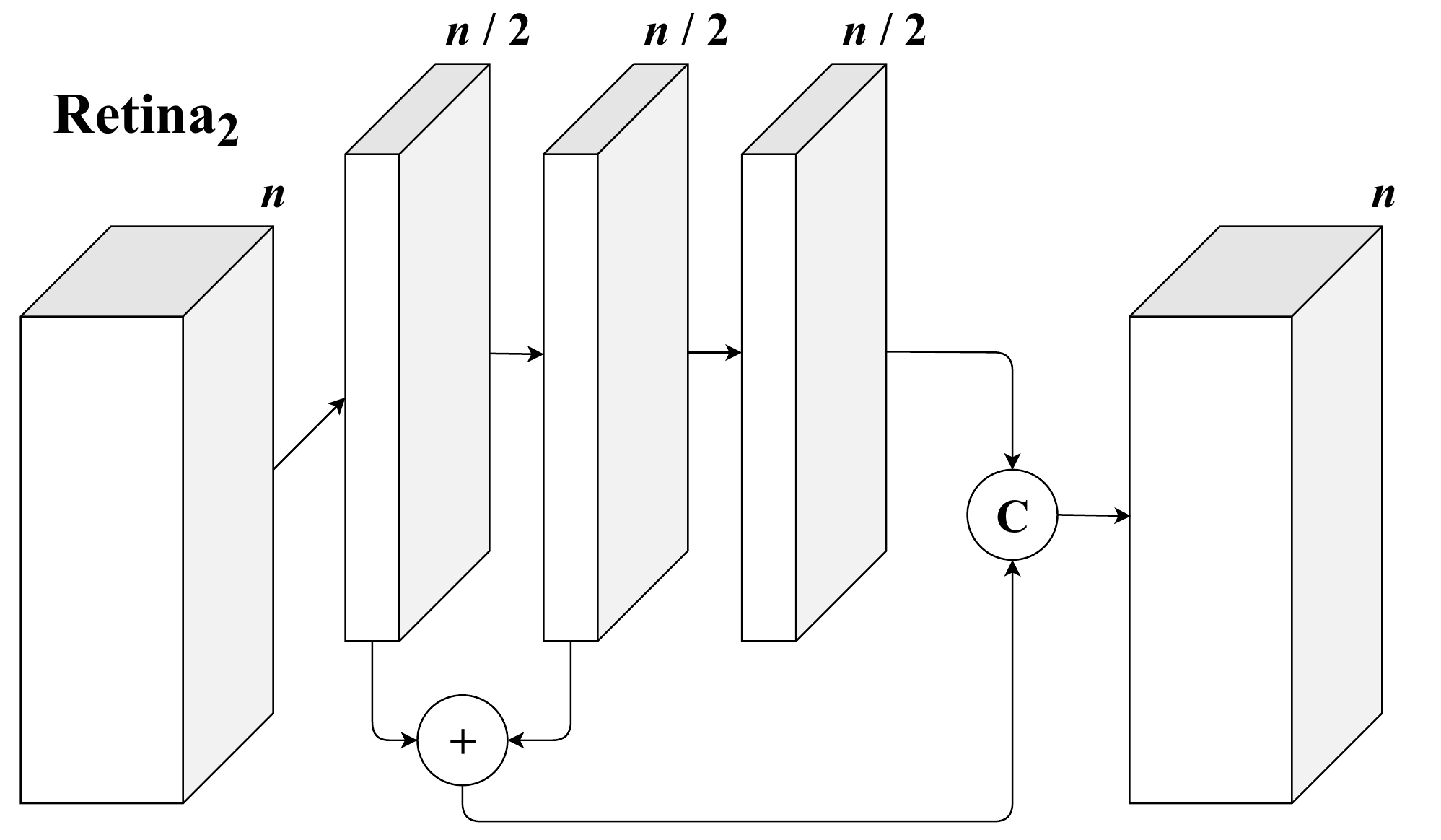}\\
\vskip 0.25cm
\includegraphics[height=0.16\textheight]{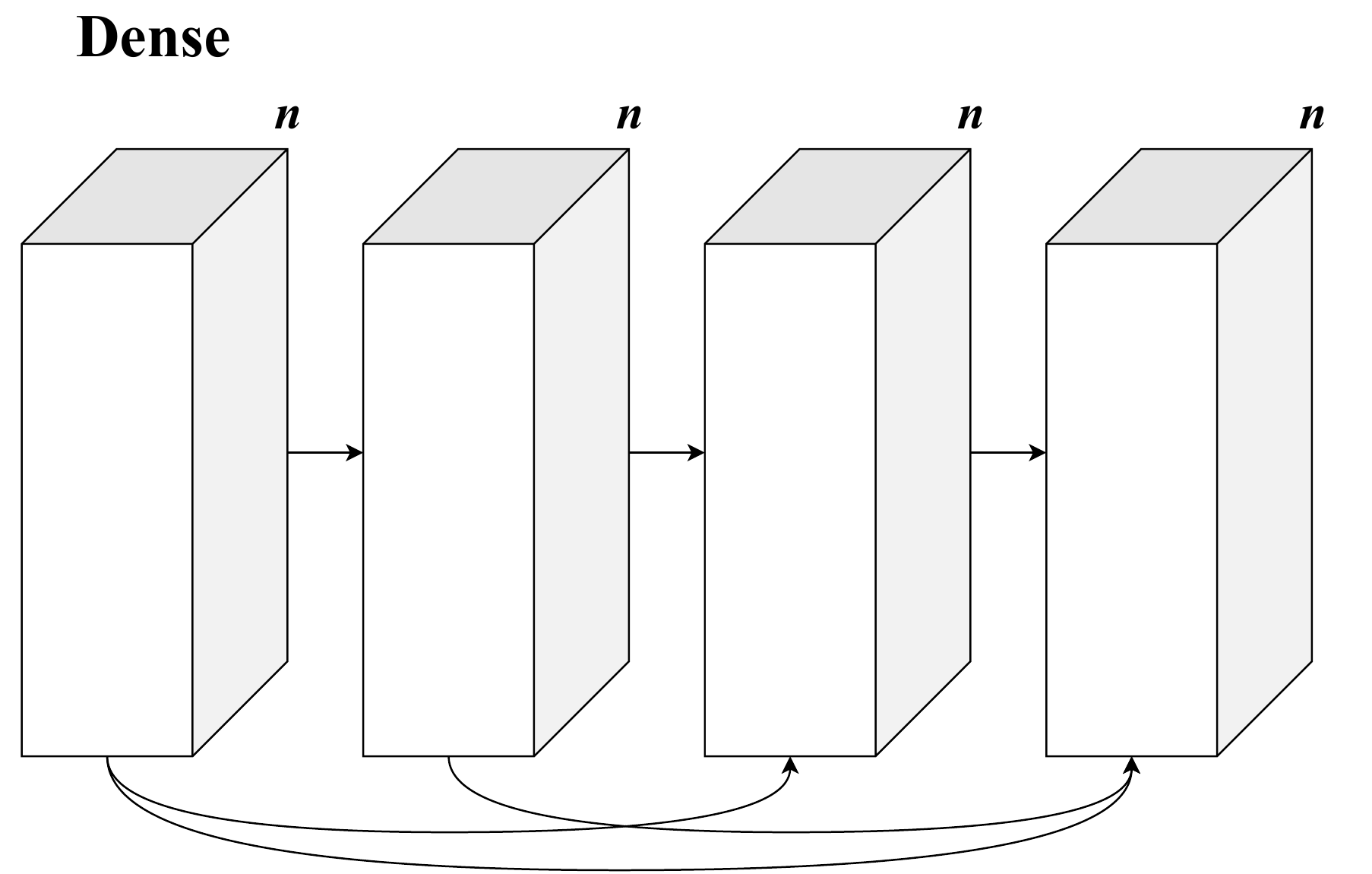}
\includegraphics[height=0.16\textheight]{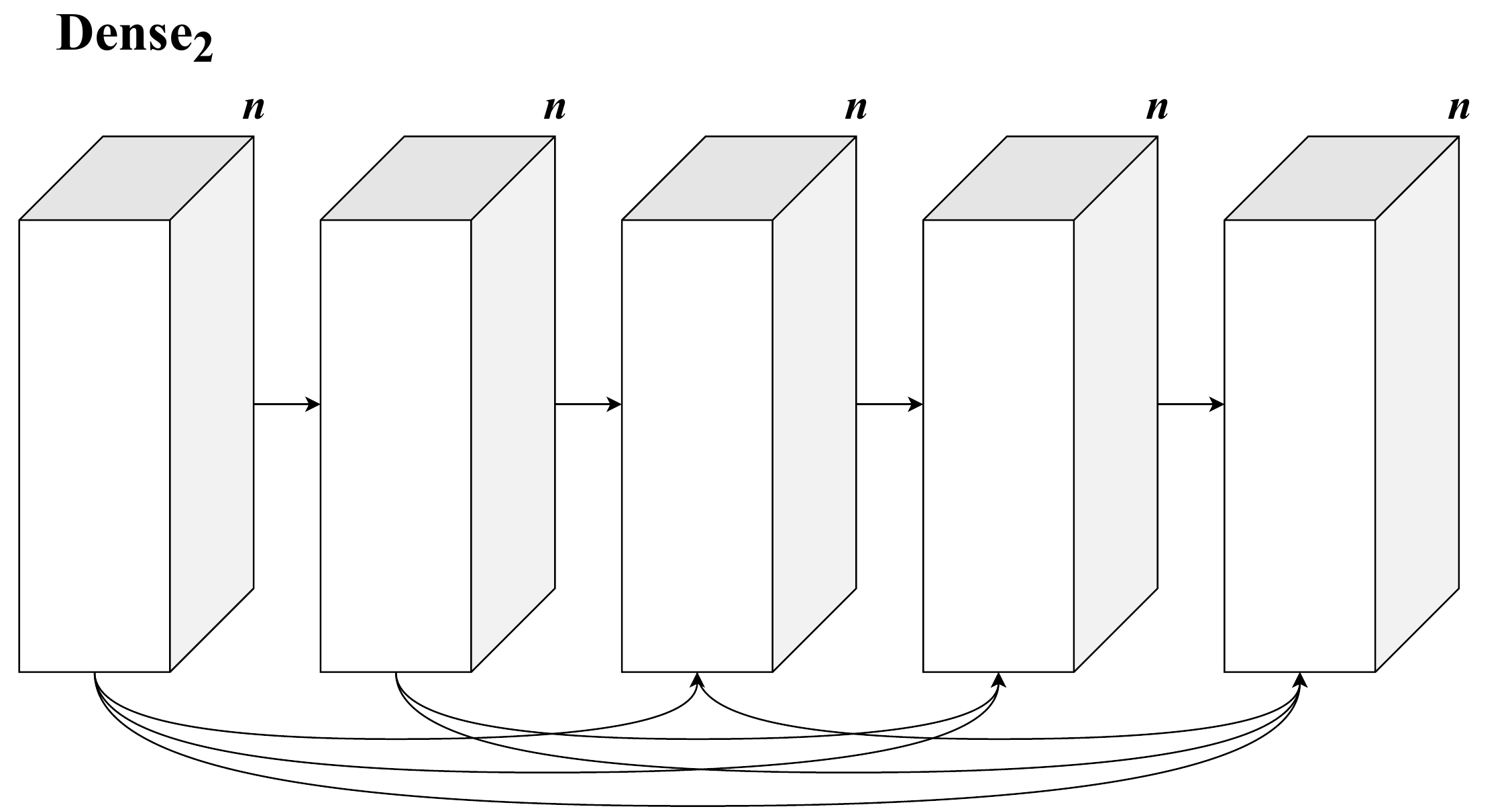}\\
\vskip 0.25cm
\caption{Illustration of the context modules used in this work, where $n$ denotes the number of filters in the first $3 \times 3$ convolution layer, `C' denotes concatenation, and `+' denotes vector addition. The left column are common context modules, SSH context module is from \cite{najibi2017}, both the RSSH and Retina modules are from \cite{deng2019}, and the Dense context module is from \cite{Li2019}. The right column are slight permutations of these modules.}
\label{fig:context}
\centering
\end{figure*}

Figure \ref{fig:context} shows an illustration of all the context modules we will be testing in our experiments. 
The left column shows the context modules from the previous section, the right column shows some slightly modified versions. 
The number of filters ($n$) in the context module is different for each network backbone.
For SSH$_2$, we modify the context module by dividing the number of channels by four then performing four $3 \times 3$ convolutions and concatenating all the outputs.
For RSSH$_2$, we simply half the number of channels in the third convolution then perform a forth $3 \times 3$ convolution and concatenate all four outputs.
For Retina$_2$, we swap the addition and concatenation, so we concatenate the last layer with the sum of the first and second layers.
For Dense$_2$, we add a fourth densely connected $3 \times 3$ convolution.
We also test with two `basic' context modules, the first with just a single $3 \times 3$ convolution and the second two concatenated $3 \times 3$ convolutions.

\section{Experiments}
\subsection{Training dataset}
To train our model we use the WIDER FACE dataset \citep{yang2016}.
This dataset consists of 32,203 images and 393,703 labeled face bounding boxes with variable scale, pose and occlusion.
The dataset is organized based on 61 event classes (e.g. parade, riot, and festival). 
Each event class is randomly sampled, with 40\%, 10\% and 50\% of the images assigned to the training, validation and testing sets.
EdgeBox \citep{zitnick2014} is used to separate the proposals into three difficulty levels; Easy, Medium and Hard with recall rates of 92\%, 76\%, and 34\%, respectively.

To incorporate landmarks into our training procedure we also use the five landmark annotations from \citep{deng2019}.
The authors labeled $\sim 84,600$ faces in the training set and made them publically available.
These landmarks follow the format used by \cite{zhang2016}: eye centers, nose tip, and mouth corners.
Faces with indistinguishable landmarks were given a dummy value and are not used in the loss function for that proposal.
\cite{deng2019} showed that by incorporating the landmarks into their multi-task loss the mAP on WIDER FACE improved by $\sim 0.7 \%$.
We also label a further $\sim 21,000$ faces in the validation set, using the same labeling scheme.


\subsection{Baseline settings}
In this report, we train with three different backbone network sizes.
We train a very lightweight network based on MobileNetV2 \citep{howard2017,sandler2018} with $\alpha = 0.25$, a heavier MobileNetV2 with $\alpha = 1.0$, and a much heavier ResNet v2 \citep{he2016} with 101 layers.
We will refer to these networks as MNet$_{\alpha=0.25}$, MNet$_{\alpha=1.0}$ and ResNet101, respectively.

\begin{tablehere}
\captionsetup{justification=centering}
\centering
\begin{tabular}{ccc}
\toprule
Pyramid & Stride & Anchor scale \\
 \midrule
$P_2$ & 4 & 16 \\
$P_3$ & 8 & 32 \\
$P_4$ & 16 & 64 \\
$P_5$ & 32 & 128 \\
$P_6$ & 64 & 256 \\
$P_7$ & 128 & 516 \\
\bottomrule
\end{tabular}
\newline
\caption{The pyramid setup we use for all of our experiments. The stride denotes the factor by which the original image have been scaled by, and the scale is the factor that is applied to the anchor size.}
\label{tab:1}
\end{tablehere}

We use an input image size of $640 \times 640$, in line with previous work \citep[e.g.][]{Li2018, deng2019} and anchor scales ranging from $16^2$ to $514^2$, with $\sim 10^5$ total anchors.
Due to the nature of the task, we set all anchors to have an aspect ratio of 1:1.
We match positive anchors with ground truth IoUs greater than $0.5$ and negative anchors with IoUs less than $0.3$.
Furthermore, we incorporate online hard example mining (OHEM) \citep{shrivastava2016} which has been successful in training other recent RPN based face detectors \citep{zhang2017,deng2019,zhang2019a,zhang2019b}.
The hard examples are selected by sorting the anchors by their loss and taking the hardest positive and negative anchors at a ratio of 1:3, following \cite{girshick2015}.
During training we randomly crop $640 \times 640$ regions of the original images \citep[following ][]{zhang2017,Tang2018,deng2019}.
For our feature pyramid we found that using a six-level feature pyramid (see Table \ref{tab:1}) gave us the best results, and we will use this setup for all our models.
We found that increasing the number of levels in the feature pyramid hampered the landmark accuracy.
However, as our primary goal is face detection accuracy, we choose to forego some landmark accuracy for improved face detection.
All of the various context modules are implemented at the same point in the network with the same tensor input all yielding the same shape output.
After each context module, we apply a modulated deformable convolution \citep{zhu2018} to enhance the context information.

Transfer learning is a widely used technique to improve the accuracy of networks and improve generalization \citep{tan2018}.
As such, all our models are pretrained on ImageNet \citep{russakovsky2014} and are finetuned on WIDER FACE.
Contrary to \cite{Zhang2019d}, we find a significant improvement in performance using transfer learning.
Our MobileNetV2 models come pretrained from {\sc glouncv} \citep{gluoncv2019}; as such our results should be reproducible.
We employ a warmup learning rate schedule \citep{goyal2017}, with five epochs where the learning rate increases linearly from $1 \times 10^{-3}$ by an order of magnitude, then falls an order of magnitude at epochs 50 and 70, and training terminates at epoch 90.
All models are trained using stochastic gradient descent with momentum $0.9$, weight decay of $5 \times 10^{-4}$ and with a batch size of eight per GPU.
The majority of our models are all trained on a single NVIDIA Telsa GPU, however, our three final models are trained across six.


\subsection{Evaluation}
WIDER FACE employs the PASCAL VOC procedure \citep{everingham2012,yang2016} for evaluation.
Detections are considered true or false based on the area of overlap with the ground truth bounding boxes.
If the intersection-over-union (IoU) between a positive anchor and ground truth is greater than $0.5$ the detection is a true positive, whereas an IoU value below this is considered a false positive.
For multiple true positive detections of one ground truth, only the detection with the highest IoU is counted as correct and the rest are counted as false positives.
The evaluation metric is average precision (AP), for each set (easy, medium, hard) the precisions are drawn from all unique recall values and averaged.

As the WIDER FACE dataset is limited to faces that are at least ten pixels high, we remove any bounding boxes with a height of fewer than five pixels.
Following \cite{najibi2017,zhang2017,li2017,wang2017face, deng2019} we employ flipping and multi-scale detection strategies, disregarding any bounding box with a class probability less than $0.02$.
We apply the greedy non-maximum suppression from \cite{girshick2013} to remove regions that have an IoU overlap greater than $0.3$ with another region that has a larger IoU with the ground truth.
We also further refine the bounding boxes by using box voting \citep{gidaris2015}, where each bounding box with an IoU overlap greater than $0.5$ `vote' on the location weighted by their respective IoU.

To evaluate the accuracy of the landmarks we use two datasets: AFW \citep[consisting of 337 faces with 68 landmarks;][]{awf} and AFLW2000 \citep[consisting of 2000 faces with 68 landmarks;][]{zhu2015}.
The defacto metric of evaluation is the mean L2 error of all the estimated landmarks normalized by the square root of the face bounding-box area (NME)---as in \citep{deng2019}.
For both datasets, we employ the same evaluation protocol, except we use the absolute (L1) error.
We calculate the absolute error (AE) using the highest confidence---center most---face, and the distance from each predicted landmarks to their respective ground truth,
\begin{equation}AE = \frac{2}{5 \sqrt{h^2 + w^2}} \sum_{i=1}^{5} | x_{i} - x_{i}^* | + | y_{i} - y_{i}^* |,\end{equation}
where $x_{i}$ and $y_{i}^*$ represent the $x$ and $y$ coordinates of the predicted and ground truth landmark, and $h$ and $w$ are the height and width of the bounding box, respectively.
This AE is then averaged over all faces in the dataset to yield the mean absolute error (mAE).
For AFLW all five landmarks are provided, however, for AFW the center of the eyes is not given.
Therefore, for AFW we use the mean of the left and right eye corner for the center of each eye.


\section{Face Detection Results}

\subsection{Backbone Baseline}
To ensure that we start with the optimal backbone we first compared the performance of all three versions of MobileNet.
Table \ref{tab:2} reports the face detection and landmark accuracies of each version of MobileNet.
We find that there is very little difference between MobileNet \citep{howard2017} and MobileNetV2 \citep{sandler2018}, however, MobileNetV3 \citep{howard2019} significantly underperforms.
As the performance of MobileNet and MobileNetV2 is so similar, we select MobileNetV2 as our backbone only because it is a lighter network.

\begin{tablehere}
\captionsetup{justification=centering}
\centering
\begin{tabular}{cccc}
\toprule
Backbone & Hard & Overall & $n_{\text{params}}$\\
\midrule
MNet$_{\alpha=0.25}$    & 87.11 & 90.90 & $8.94 \times 10^5$ \\
MNetV2$_{\alpha=0.25}$   & 87.27 & 91.03 & $7.08 \times 10^5$\\
MNetV3-Small             & 86.52 & 89.37 & $1.46 \times 10^6$ \\
\midrule
MNet$_{\alpha=1.0}$     & 89.10 & 93.10 & $4.57 \times 10^6$ \\
MNetV2$_{\alpha=1.0}$    & 89.87 & 93.28 & $3.62 \times 10^6$ \\
MNetV3-Large             & 88.23 & 92.12 & $4.33 \times 10^6$ \\
\bottomrule
\end{tabular}
\newline
\caption{Generic mobilnet backbones and their respective performance on WIDER FACE. The left column is the performance on the `hard' set, the middle is the performance averaged across all three sets, and the right column is the number of parameters}
\label{tab:2}
\end{tablehere}

For the rest of this paper, we will be referring to three network backbones.
Our smallest network (MNet$_{\alpha=0.25}$) we use a MobileNetV2 backbone with an $\alpha$ value of $0.25$ and just $n = 64$ filters in the context modules. 
The medium sized network (MNet$_{\alpha=1.0}$) also uses a MobileNetV2 backbone with $\alpha = 1.0$ and $n = 128$ filters in the context modules. 
For our large network, we choose ResNet101 v2 with $n = 256$ filters in the context modules to be comparable with other literature.

\subsection{Context Module Comparison: Face Detection}

\begin{table*}
\captionsetup{justification=centering}
\centering
\begin{tabular}{lcccccc}
\toprule
 & \multicolumn{2}{c}{MNet$_{\alpha=0.25}$ (AP \%)} & \multicolumn{2}{c}{MNet$_{\alpha=1.0}$ (AP \%)} & \multicolumn{2}{c}{ResNet101 (AP \%)} \\
Head & Hard ($\Delta$) & Overall ($\Delta$) & Hard ($\Delta$) & Overall ($\Delta$) & Hard ($\Delta$) & Overall ($\Delta$) \\
\midrule
SSH             & 86.85 (+0.15) & 90.79 (-0.01) & 89.04 (+0.11) & 92.76 (+0.14) & 90.55 (-0.08) & 94.14 (+0.05) \\
SSH$_2$         & 86.76 (+0.06) & 90.86 (+0.05) & 88.76 (-0.17) & 92.58 (-0.05) & 90.64 (+0.01) & 94.22 (+0.13) \\
Retina          & 87.11 (+0.42) & 91.02 (+0.21) & 88.87 (-0.06) & 92.59 (-0.04) & 90.74 (+0.11) & 94.26 (+0.24) \\
Retina$_2$      & 87.27 (+0.57) & 91.03 (+0.22) & 88.70 (-0.23) & 92.43 (-0.19) & 90.49 (-0.14) & 94.21 (-0.01) \\
RSSH            & 86.41 (-0.29) & 90.67 (-0.14) & 89.19 (+0.26) & 92.69 (+0.06) & 90.68 (+0.05) & 94.18 (+0.18) \\
RSSH$_2$        & 86.67 (-0.03) & 90.83 (+0.02) & 88.99 (+0.06) & 92.61 (-0.01) & 90.68 (+0.05) & 94.25 (+0.17) \\
Dense           & 86.47 (-0.23) & 90.56 (-0.25) & 88.93 (+0.00) & 92.57 (-0.05) & 90.52 (+0.00) & 94.21 (0.02) \\
Dense$_2$       & 86.05 (-0.65) & 90.70 (-0.11) & 88.97 (+0.04) & 92.77 (+0.14) & 89.72 (+0.00) & 93.30 (-0.77) \\
\midrule
Average         & 86.70 $\pm$ 0.39 & 90.81 $\pm$ 0.16 & 88.93 $\pm$ 0.16 & 92.63 $\pm$ 0.11 & 90.50 $\pm$ 0.33 & 94.07 $\pm$ 0.32 \\
\bottomrule
\end{tabular}
\newline
\caption{The WIDER FACE validation average precision (AP) for each backbone and each context module. The value in parentheses denotes the APs divergence ($\Delta$) from the mean. The `Hard' columns show the APs on the `hard' set, and the `Overall' columns show the mean APs across all three sets. The bottom row shows the mean APs and standard deviation for all context modules.}
\label{tab:ctx_wider}
\end{table*}

Table \ref{tab:ctx_wider} presents the results for each context module on the WIDER FACER validation dataset.
For reference we also trained MNetV2$_{\alpha = 0.25}$ with two `basic' context modules, our two layer module only achieves $85.15\%$ and $89.94\%$, and our one layer module only $85.15\%$ and $89.52\%$ on the `hard' set and overall, respectively.
Therefore, we can see that having at least three layers improved the performance by more than $1$ percent.

For MNetV2$_{\alpha=0.25}$, we find that Retina$_2$, Retina, and SSH are the three top performers on the `hard' set, and Retina$_2$, Retina, and SSH$_2$ the top overall.
For MNetV2$_{\alpha=1.0}$, we find that Retina$_2$, Retina, and RSSH$_2$ are the three top performers on the `hard' set, and SSH, Dense$_2$, and RSSH the top overall.
For ResNet101, we find that Retina, RSSH, and RSSH$_2$ are the three top performers on the `hard' set, and Retina, RSSH$_2$, and SSH$_2$ are the top overall.
For the `hard' set, we find that the top three context performers are Retina, Retina$_2$ and SSH with average mean divergences of $0.16\%$, $0.07\%$ and  $0.06\%$, respectively.
Over all three sets, we find that the top three context performers are Retina, SSH and RSSH$_2$ with average mean divergences of $0.08\%$, $0.02\%$ and  $0.02\%$, respectively.
From these results the context modules seem quite similar in performance, the only real outliers are Dense and Dense$_2$ which seem to underperform.

To test the statistical significance of these results we ran the same experiment eight times (to be the same number as the number of context modules) to determine the amount of variance just due to randomness.
We find almost no difference in variance between running the same experiment and using different context modules.
For the `hard' set we get standard deviations of $0.35$, $0.11$ and $0.07$ for  MNetV2$_{\alpha=0.25}$, MNetV2$_{\alpha=1.0}$ and ResNet101, respectively.
Overall three sets we get standard deviations of $0.15$, $0.11$ and $0.05$ for  MNetV2$_{\alpha=0.25}$, MNetV2$_{\alpha=1.0}$ and ResNet101, respectively.
Therefore, we can not find any significant difference between any of the architectures.
We can also see that smaller networks have higher variance.
Therefore, it is unlikely that the architecture of the context module has more importance for smaller networks, it is just that the randomness is more influential.
We also compared the number of filters used in the context module, finding on average a $\sim 1$ percent performance increase for smaller networks when doubling the number of filters.
However, this comes with significant efficiency problems.
For example, doubling the number of filters in the context modules for MNet$_{\alpha=0.25}$ almost doubles the total amount of parameters in the network.


\subsection{Context Module Comparison: Landmark Accuracy}

\begin{table*}
\captionsetup{justification=centering}
\centering
\begin{tabular}{lcccccc}
\toprule
 & \multicolumn{2}{c}{MNetV2$_{\alpha=0.25}$ (mAE $10^{-2}$)} & \multicolumn{2}{c}{MNetV2$_{\alpha=1.0}$ (mAE $10^{-2}$)} & \multicolumn{2}{c}{ResNet101 (mAE $10^{-2}$)} \\
Head & AFW & AFLW & AFW & AFLW & AFW & AFLW \\
\midrule
SSH             & 1.14 $\pm$ 0.62 & 1.86 $\pm$ 2.80 & 1.03 $\pm$ 0.56 & 1.60 $\pm$ 1.92 & 0.91 $\pm$ 0.53 & 1.56 $\pm$ 1.83 \\
SSH$_2$         & 1.55 $\pm$ 2.76 & 2.05 $\pm$ 2.42 & 0.98 $\pm$ 0.98 & 1.51 $\pm$ 1.95 & 0.92 $\pm$ 0.54 & 1.51 $\pm$ 2.50 \\
Retina          & 1.12 $\pm$ 0.63 & 1.84 $\pm$ 2.68 & 0.95 $\pm$ 0.59 & 1.63 $\pm$ 2.20 & 0.90 $\pm$ 0.51 & 1.41 $\pm$ 1.47 \\
Retina$_2$      & 1.09 $\pm$ 0.68 & 2.00 $\pm$ 3.15 & 1.02 $\pm$ 0.57 & 1.61 $\pm$ 1.95 & 1.00 $\pm$ 0.87 & 1.63 $\pm$ 2.24 \\
RSSH            & 1.11 $\pm$ 0.90 & 1.99 $\pm$ 3.09 & 1.01 $\pm$ 0.52 & 1.42 $\pm$ 1.85 & 0.90 $\pm$ 0.55 & 1.50 $\pm$ 1.99 \\
RSSH$_2$        & 1.12 $\pm$ 1.31 & 2.14 $\pm$ 3.33 & 1.01 $\pm$ 0.99 & 1.54 $\pm$ 2.74 & 0.97 $\pm$ 0.73 & 1.53 $\pm$ 2.06 \\
Dense           & 1.15 $\pm$ 2.78 & 2.11 $\pm$ 3.16 & 1.21 $\pm$ 0.66 & 1.58 $\pm$ 2.00 & 0.93 $\pm$ 0.62 & 1.53 $\pm$ 2.02 \\
Dense$_2$       & 1.27 $\pm$ 2.63 & 2.00 $\pm$ 2.92 & 1.05 $\pm$ 1.39 & 1.48 $\pm$ 2.07 & 0.93 $\pm$ 0.62 & 1.53 $\pm$ 2.02 \\
\midrule
Average         & 1.19 $\pm$ 1.80 & 2.00 $\pm$ 2.96 & 1.03 $\pm$ 0.78 & 1.55 $\pm$ 2.10 & 0.93 $\pm$ 0.62 & 1.53 $\pm$ 2.02 \\
\bottomrule
\end{tabular}
\newline
\caption{The mean absolute error (mAE) of the predicted landmarks, normalized by the bound box size, for each backbone and context module. The bottom row shows the average mAE and the pooled standard deviation across all context modules. We compare our five landmarks to the five closest in both datasets, however, for AFW we use the average coordinates of the eye corners for each eye. We can see a significant drop in error by increasing the model size, but no significant difference in context module choice.}
\label{tab:ctx_lmk}
\end{table*}

Table \ref{tab:ctx_lmk} presents the mean absolute error and standard deviation for each backbone and context module.
Similarly to the previous section, we also present the accuracy of two MNetV2$_{\alpha = 0.25}$ networks trained with `basic' context modules. 
Our single layer context module network can achieve $1.24 \pm 2.03 \times 10^{-2}$ and $2.22 \pm 3.30 \times 10^{-2}$ on AFW and ALFW-2000, respectively.
Moreover, our two layer context module network can achieve $1.27 \pm 1.12 \times 10^{-2}$ and $2.17 \pm 3.14 \times 10^{-2}$ on AFW and ALFW-2000, respectively.
These results are not far from the accuracy of the other context modules presented in Table \ref{tab:ctx_lmk}, therefore, the choice of context module seems to have little influence on landmark accuracy.
Following the previous section, to investigate the effect of randomness we compare these results to a single experiment run eight times.
When switching context modules we get $\sigma = 0.15$, $\sigma = 0.08$ and $\sigma = 0.04$ for MNet$_{\alpha=0.25}$, MNet$_{\alpha=1.0}$ and ResNet101, respectively.
But when running the same experiment eight times we get $\sigma = 0.002$, $\sigma = 0.0012$ and $\sigma = 0.0004$ for MNet$_{\alpha=0.25}$, MNet$_{\alpha=1.0}$ and ResNet101, respectively.
As in the previous section, we find that the variance due to randomness is similar to the variance in the choice of the context module architecture.

For MNetV2$_{\alpha=0.25}$, we find average mean absolute errors of $1.19 \pm 1.80 \times 10^{-2}$ and $2.0 \pm 2.96 \times 10^{-2}$ for AFW and AFLW-2000, respectively.
Whereas, MNetV2$_{\alpha=1.0}$ is significantly better on both datasets, with an average mean absolute error of $1.03 \pm 0.78 \times 10^{-2}$ and $1.55 \pm 2.10 \times 10^{-2}$ for AFW and AFLW-2000, respectively.
Furthermore, the largest backbone, ResNet101, is even better, with an average mean absolute error of $0.93 \pm 0.62 \times 10^{-2}$ and $1.53 \pm 2.02 \times 10^{-2}$ for AFW and AFLW-2000, respectively.
Unsurprisingly, the larger backbones perform significantly better on landmark localization, especially on side faces.

We also consider the impact of increasing the number of filters in the context module on the landmark accuracy.
We find that doubling the number of filters in the context module has almost no effect on the quality of the landmarks.
These results suggest the choice of context module arcutecture, and the number of filters it has is irrelevant to landmark quality.


\subsection{Final Results: Face Detection}

\begin{tablehere}
\captionsetup{justification=centering}
\centering
\begin{tabular}{cccc}
\toprule
 MNetV2$_{\alpha = 0.25}$ & MNetV2$_{\alpha = 1.0}$ & ResNet152 \\
 \midrule
$7.1 \times 10^{5}$ & $4.5 \times 10^{6}$ & $4.5 \times 10^{7}$\\
\bottomrule
\end{tabular}
\newline
\caption{Total number of parameters for each backbone.}
\label{tab:finalparamas}
\end{tablehere}

\begin{tablehere}
\captionsetup{justification=centering}
\centering
\begin{tabular}{lcccc}
\toprule
Backbone & Hard (AP\%) & Overall (AP\%) \\
 \midrule
MNetV2$_{\alpha = 0.25}$    & 87.37 & 91.49 \\
MNetV2$_{\alpha = 1.0}$     & 90.16 & 93.60 \\
ResNet152                   & 91.71 & 94.78 \\
\bottomrule
\end{tabular}
\newline
\caption{Average precision of each backbone on the WIER FACE `hard' set and averaged over all sets. All models are trained with a batch size of eight across six GPUs and are pretrained on ImageNet \citep{russakovsky2014}}
\label{tab:finalwider}
\end{tablehere}

For our final models we only change the number of GPUs (from one to six) and the number of layers in the ResNet v2 (from 101 to 152).
Table \ref{tab:finalwider} shows our final results on the WIDER FACE `hard' set and overall sets.
We find that increasing the size of the network is the most reliable way to increase the performance on the WIDER FACE dataset.
However, using networks like ResNet152 may not be practical in most applications.
Both of our MobileNetV2s perform extremely well.
For comparison, \cite{zhang2019b} presents results for their ResNet18 model which achieves very similar accuracy to our much smaller MNetV2$_{\alpha=1.0}$ both achieving $\sim 90.2 \%$ on the `hard' set.
Moreover, \cite{deng2019} report their results on the `hard' set using MobileNet with $\alpha = 0.25$, with an AP of just $78.2\%$, whereas, our model with the same backbone achieves $87.4\%$ using a six-layer pyramid and $85.3\%$ using only three-layers.
By increasing the number of filters in the context module by a factor of two we can achieve over $88\%$ on the `hard' set.
However, this makes the network significantly bigger, so it is not a fair comparison with \cite{deng2019}.

\begin{figure*}[t!]
\centering
\includegraphics[width=0.99\textwidth]{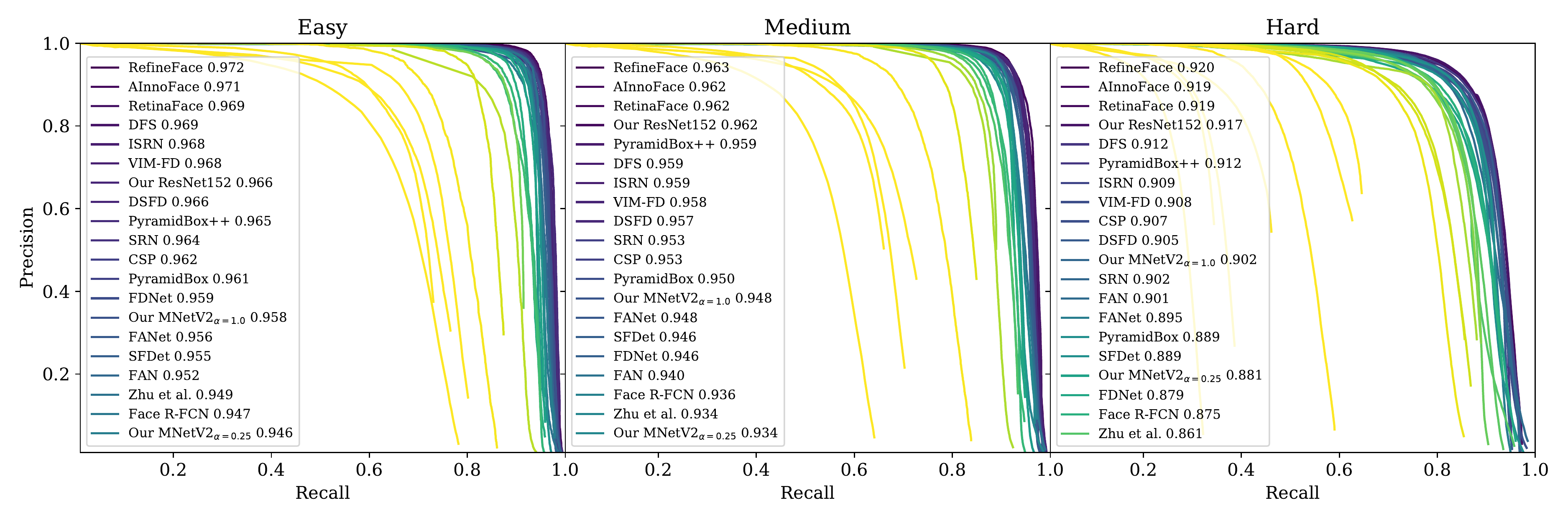}
\caption{Final results for face detection on the WIDER FACE validation set. In the figure we show the curves for all submissions but only label the top 20.}
\label{fig:wider}
\end{figure*}

Figure \ref{fig:wider} shows our final results on all three of the WIDER FACE validation sets.
Our ResNet152 can rank a respectable fourth on both the `medium' and `hard' sets, without adding a large number of layers.
Moreover, our MNetV2$_{\alpha = 0.25}$ can rank four to five places higher than EXTD (a similar lightweight detector) on all three sets.
Also, our MNetV2$_{\alpha = 1.0}$ is able to out perform much heavier networks, for exmaple FAN \citep{zhang2019a} which uses a ResNet50 backbone.

\subsection{Final Results: Landmark Accuracy}

\begin{tablehere}
\captionsetup{justification=centering}
\centering
\begin{tabular}{lcccc}
\toprule
                            & AFW              & AFLW \\
Backbone                    & (mAE$10^{-2}$)   & (mAE$10^{-2}$) \\
\midrule
MNetV2$_{\alpha = 0.25}$    & 1.06 $\pm$ 1.11 & 1.80 $\pm$ 2.39 \\
MNetV2$_{\alpha = 1.0}$     & 0.79 $\pm$ 0.43 & 1.17 $\pm$ 1.42 \\
ResNet152                   & 0.87 $\pm$ 1.52 & 0.87 $\pm$ 0.67 \\
\bottomrule
\end{tabular}
\newline
\caption{Mean absolute error ($10^{-2}$) of the landmark predictions for each backbone. We can see that heavier models perform much better on landmark localization.}
\label{tab:finallmk}
\end{tablehere}

Table \ref{tab:finallmk} shows the final landmark accuracy for each of our backbones and table \ref{tab:finalparamas} shows the total number of parameters in the network.
For comparison MTCNN \citep{zhang2016} achieves $\text{mAE} = 3.64 \pm 2.232 \times 10^{-2}$ on AFLW-2000, far higher than even our smallest model, and struggles to even detect many of the faces in AFW.
We can see that larger backbones, generally, provide better quality landmarks.
As previously mentioned, landmark accuracy can be improved significantly by reducing the number of layers in the feature pyramid.
However, this will cause a substantial loss in face detection accuracy.
We also investigated including more filters in the context modules, which does not affect the quality of landmarks.
Moreover, we found that using deformable convolutions also impeded the landmark accuracy, but again it is a trade-off we make to ensure higher face detection accuracy.

\subsection{Final Results: Network Performance}

Table \ref{tab:finalperf} shows the inference speed of both MobileNetV2 models on different devices. 
These devices are very heterogeneous, so we use specialized techniques to get the best performance from each of them.
All of the optimization techniques are open source, therefore, these benchmarks should be reproducible.
For desktop CPU (Intel i5-7500) and embedded GPU (NVIDIA Jetson TX2) we take advantage of {\sc tvm} \citep{tvm}.
While, {\sc mxnet} \citep{MXNet} accelerated using {\sc cuda} \citep{cuda} is used for the desktop GPU (NVIDIA 1050ti).

\begin{tablehere}
\captionsetup{justification=centering}
\centering
\begin{tabular}{lccccccc}
\toprule
Device & \multicolumn{2}{c}{MNetV2$_{\alpha=0.25}$} & \multicolumn{2}{c}{MNetV2$_{\alpha=1.0}$} \\
 & ms & fps & ms & fps \\
\midrule
Desktop CPU     & 60.3 & 16.6 & 142.2 & 7.0 \\
Mobile CPU      & 38.1 & 26.2 & 89.3 & 11.2 \\
Embedded GPU    & 34.3 & 29.2 & 77.3 & 12.9 \\
Mobile GPU      & 17.6 & 56.8 & 39.9 & 25.1 \\
Desktop GPU     & 9.1 & 109.9 & 20.9 & 47.8 \\
\bottomrule
\end{tabular}
\newline
\caption{For for the desktop we use one core of an intel i5-7500 (CPU), and an NVIDIA 1050ti (GPU). For the mobile device we use a Xiaomi Mi9 with a SnapDragon 855 chipset (Kryo 485 CPU and Adreno 640 GPU). While, the embedded GPU is a NVIDIA Jetson TX2.}
\label{tab:finalperf}
\end{tablehere}

For comparison, EXTD \citep{yoo2019}, use far less parameters in their face detector; just $\sim 10^{5}$.
However, our smallest model is not only $34 - 41$ ms faster on a VGA input, it also performs better on the WIDER FACE hard set by $1.8\% - 4.9\%$.
RefineFace \citep{zhang2019b} use ResNet-18 as their smallest model which runs in 26.8ms on an NVIDIA 1080ti.
By comparison, we achieve similar results on the WIDER FACE hard set with our MNetV2$_{\alpha = 1.0}$, which runs $\sim6$ms faster on a much slower GPU (NVIDIA 1050ti).
To compare to \cite{deng2019}, we use a three-layer pyramid and can achieve very similar or better inference speeds. 
We also achieve a higher accuracy on WIDER FACE hard set with the same model.


\section{Conclusions}
We have shown that the choice of context module architecture is likely irrelevant to the models' performance.
One possible reason for this is that the layers added by the feature pyramid and the context modules are always randomly initialized and, for smaller networks, they can constitute a large number of the total parameters $\sim 25$ percent.
Therefore, a `lucky' initialization can yield more performance gain than crafting an optimal context module.
One possible way around this would be to pretrain the full network on a similar detection task, e.g. person detection, to alleviate the effect of the random initialization.

Our largest model can achieve a near state-of-the-art score on the WIDER FACE hard set of $\sim 91.7$ percent without making use of any excessive additional layers. 
It also provides very accurate landmarks that can be used for face alignment.
Our two smaller networks can exceed state-of-the-art performance on the WIDER FACE hard set compared to similar network sizes.
These networks also provide accurate landmarks while being able to run in real-time on modest desktop and mobile hardware.


\section*{Acknowledgements}
We would like to thank Aubin Samacoits, Jeff Hnybida, Riccardo Gallina and Sanjana Jain for their constructive input and feedback during the writing of this paper.
We would also like to thank CAT Telecom for granting us access to their GPU cluster for training.


\bibliographystyle{unsrt}  
\bibliography{references}

\end{multicols}

\end{document}